\documentclass[
]{ceurart}

\usepackage{listings}
\usepackage{cleveref}
\usepackage{tikz}
\usetikzlibrary{arrows.meta, positioning, shapes.geometric}
\usepackage{listings}
\usepackage{float}
\usepackage{subcaption}

\usepackage[most]{tcolorbox}
\tcbset{
  colback=gray!5!white, 
  colframe=gray!50!black, 
  colbacktitle=black,
  coltitle=white,
  fonttitle=\bfseries\footnotesize,
  boxrule=0.5pt, 
  arc=3pt, 
  left=6pt, 
  right=6pt, 
  top=2pt, 
  bottom=2pt, 
  boxsep=4pt,
  enhanced,
}

\begin{document}

%%
%% Rights management information.
%% CC-BY is default license.
\copyrightyear{2025}
\copyrightclause{Copyright for this paper by its authors.
  Use permitted under Creative Commons License Attribution 4.0
  International (CC BY 4.0).}

%%
%% This command is for the conference information
% \conference{ISWC'25: International Semantic Web Conference, Nara, Japan}
\conference{RAGE-KG 2025: The Second International Workshop on Retrieval-Augmented Generation Enabled by Knowledge Graphs, co-located with ISWC 2025, November 2--6, 2025, Nara, Japan}

%%
%% The "title" command
% \title{RAG-FLARKO: Enhancing Context for Financial Recommendations via Retrieval-Augmented Knowledge Graphs}
% \title{RAG-FLARKO: Leveraging Retrieval-Augmented Knowledge Graphs for Personalized Financial Recommendations}
% \title{A Knowledge Graph-Enhanced Retrieval-Augmented Generation Framework for Financial Recommendations}
\title{Parallel and Multi-Stage Knowledge Graph Retrieval for Behaviorally Aligned Financial Asset Recommendations}

%%
%% The "author" command and its associated commands are used to define
%% the authors and their affiliations.
\author[1]{Fernando Spadea}[%
orcid=0009-0006-4278-3666,
email=spadef@rpi.edu,
% url=https://yamadharma.github.io/,
]
\cormark[1]
\address[1]{Rensselaer Polytechnic Institute, Troy, NY, USA}

\author[1]{Oshani Seneviratne}[%
orcid=0000-0001-8518-917X,
email=senevo@rpi.edu,
url=https://oshani.info,
]

%% Footnotes
\cortext[1]{Corresponding author.}

%%
%% The abstract is a short summary of the work to be presented in the
%% article.
\begin{abstract}
Large language models (LLMs) show promise for personalized financial recommendations but are hampered by context limits, hallucinations, and a lack of behavioral grounding. 
Our prior work, FLARKO, embedded structured knowledge graphs (KGs) in LLM prompts to align advice with user behavior and market data. This paper introduces RAG-FLARKO, a retrieval-augmented extension to FLARKO, that overcomes scalability and relevance challenges using multi-stage and parallel KG retrieval processes. Our method first retrieves behaviorally relevant entities from a user's transaction KG and then uses this context to filter temporally consistent signals from a market KG, constructing a compact, grounded subgraph for the LLM. This pipeline reduces context overhead and sharpens the model's focus on relevant information. Empirical evaluation on a real-world financial transaction dataset demonstrates that RAG-FLARKO significantly enhances recommendation quality. Notably, our framework enables smaller, more efficient models to achieve high performance in both profitability and behavioral alignment, presenting a viable path for deploying grounded financial AI in resource-constrained environments.
\end{abstract}

%%
%% Keywords. The author(s) should pick words that accurately describe
%% the work being presented. Separate the keywords with commas.

%%
%% This command processes the author and affiliation and title
%% information and builds the first part of the formatted document.
\maketitle

\section{Introduction}

The task of \textbf{financial asset recommendation}, i.e., suggesting assets such as stocks or bonds to investors, is a knowledge-intensive process that requires personalization, transparency, and factual accuracy. While Large Language Models (LLMs) are powerful tools for generating financial advice, they face key limitations, including limited context windows and susceptibility to hallucinations. A critical challenge is their weak \textbf{behavioral grounding}: ensuring that recommendations align with an investor's past behavior and preferences, as reflected in their transaction history. An emerging question is whether generative AI can provide trusted, customized financial advice. Central to this is the need to ground generated advice in verifiable evidence, including market data or historical trends, rather than merely plausible-sounding text. Retrieval-augmented generation (RAG) offers a promising solution by enabling models to retrieve and incorporate real-time, external information into the recommendation process.

To address these challenges, our previous work on FLARKO (Financial Language-model for Asset Recommendation with Knowledge-graph Optimization)~\cite{rag-flarko}, introduced a framework that grounds LLM reasoning using two complementary knowledge graphs: a personal KG (PKG), capturing user transaction histories, and a market KG (MKG), representing market-level asset performance. The model is fine-tuned with Kahneman-Tversky Optimization~\cite{ethayarajh2024kto} due to its performance under both centralized and federated settings~\cite{spadea2025federated}. The project's GitHub repository~\cite{rag-flarko} contains the full documentation for the entire framework, including its centralized and federated implementations (CenFLARKO and FedFLARKO), as well as extensive evaluations against a range of baseline models for financial asset recommendations. 

\begin{figure}[t]
    \centering
    \includegraphics[width=\linewidth]{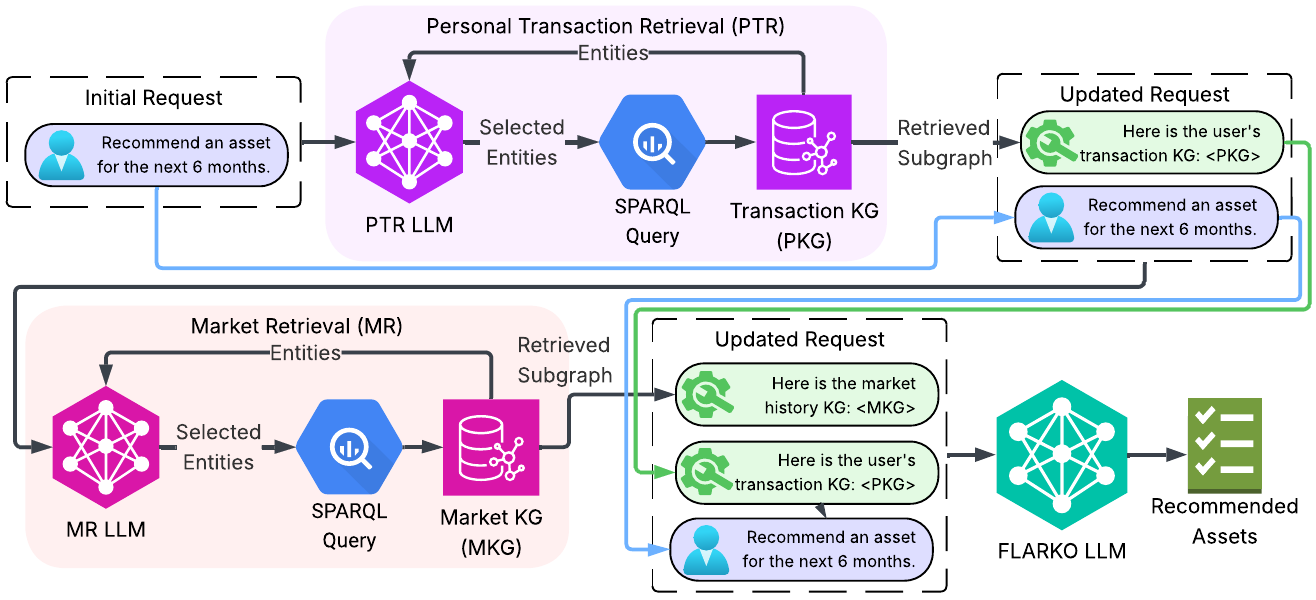}
    % \captionsetup{justification=centering} 
    \caption{RAG-FLARKO Multi-Stage Retrieval Pipeline}
    \label{fig:pipe}
\end{figure}

While effective, the original FLARKO approach of full KG injection into the LLM context incurs significant token costs and struggles to scale to larger KGs or more complex queries.
To address these limitations, in this paper, we introduce \textbf{RAG-FLARKO}, an extension of FLARKO that combines retrieval-augmented generation with structured KG reasoning.
This approach enables more efficient “needle-in-a-haystack” retrieval in financial recommendation tasks by selectively injecting only the most relevant context. By decomposing KG inclusion into two targeted retrieval steps, RAG-FLARKO assembles a minimal, temporally filtered subgraph tailored to each user request. This design maintains behavioral grounding while addressing the scalability and context limitations of traditional KG injection directly into the LLM context.
As shown in \Cref{fig:pipe}, first, a \emph{Personal Transaction Retrieval} (PTR) stage issues creates a relevant subgraph from a KG of the user's transaction history. Next, a \emph{Market Retrieval} (MR) stage creates a second relevant subgraph from a KG of the overall market, using the contextual information retrieved in the previous PTR stage. These subgraphs are then provided to the FLARKO model for responding to the user's request. By decomposing retrieval with temporal filtering and leveraging KG structure, our pipeline overcomes LLM context limits and prevents data leakage, ensuring that all recommendations are grounded in facts available at query time.

\subsection*{Our Contributions:}
\begin{itemize}
    \item \textbf{Multi-step, KG-driven retrieval:} A two-stage RAG pipeline that issues SPARQL queries over user transaction and market KGs to chain behavior to relevant financial signals.
    
    \item \textbf{LLM-based entity selection and subgraph construction:} We apply LLMs to filter relevant entities and construct compact, context-efficient KG subgraphs for recommendation generation.
    
    % \item \textbf{Temporal-aware KG filtering:} All retrievals enforce a recommendation-date cutoff, ensuring that generated advice reflects only the information available at inference time.
    
    % OS: TODO -- this contribution is to be verified
    \item \textbf{Context-window optimization via RAG-FLARKO:} Our method reduces token footprint, enabling smaller LLMs to outperform full-KG models in behavioral alignment and profitability.
    
    \item \textbf{Empirical evaluation in financial recommendation:} We evaluate RAG-FLARKO on the FAR-Trans dataset~\cite{sanz2024far}, demonstrating improved recommendation quality over baseline FLARKO in terms of both profitability and behavioral alignment.
\end{itemize}

% \begin{itemize}
%     \item Limitations in context size in FLARKO fixed with RAG-FLARKO
%     \item Two-stage retrieval process for personalized financial recommendations: user's past transaction history is used in contextual chaining where user behavior informs market data relevance
% \end{itemize}

% OS: FL stuff maybe, if we have the time

% OS: I moved the figure up

\section{Related Work}

\subsection{KGs in Financial Recommendation Systems}

KGs have become a core tool for enhancing recommender systems, providing relational structure and semantic context that go beyond user–item interaction matrices~\cite{guo2020survey}.
In financial applications, where relationships between investors, instruments, markets, and institutions are intricate and dynamic, KGs enable more explainable and adaptable recommendations~\cite{sun2025knowledge}.
In consumer-facing applications such as robo-advisors, KGs help link investor preferences with product attributes to produce tailored suggestions~\cite{shen2025artificial}. 
In financial news recommendation, KGs help encode entities and events to contextualize article relevance based on user portfolios or interests~\cite{ren2019financial}. 
This illustrates a broader trend: KGs enhance personalization and interpretability by embedding domain knowledge into the recommendation pipeline.

Several systems further demonstrate the utility of KGs:
\begin{itemize}
    \item FinDKG~\cite{li2024findkg} constructs a dynamic knowledge graph from real-time financial news, capturing evolving sector trends for thematic investment strategies. While powerful for macroeconomic reasoning, FinDKG is not tailored to individual user behavior scenarios like ours.
    \item \citet{tang2023intelligent} present a stock recommender based on a generalized financial KG representing companies, industries, and market signals. Their model supports thematic matching via graph traversal but does not support behavior-specific filtering or LLM-based recommendation generation.
    \item \citet{verma2023empowering} introduce an interpretable article recommender that builds a KG from structured and unstructured data, supporting both XGBoost and reinforcement learning-based inference. While they highlight the value of KG traversal paths, their system focuses on content recommendation rather than asset allocation or behavioral alignment.
    \item The FNRKPL framework~\cite{sun2025knowledge} translates KG triples into natural language prompts to guide news recommendation. This approach injects KG-derived facts into LLM prompts but lacks structured subgraph retrieval and personalization, two core features of our RAG-FLARKO pipeline.
\end{itemize}

In contrast to these works, our method targets behaviorally aligned financial asset recommendation using structured personal transaction and market KGs, and introduces a multi-stage retrieval process to construct compact, temporally valid, user-specific KG subgraphs.

\subsection{RAG for Financial Tasks}

Originally proposed by \citet{lewis2020retrieval}, RAG combines a retriever with a generative language model, enabling outputs that are grounded in external information. RAG has proven effective in factual domains such as law and medicine, and is increasingly being adopted for financial tasks.
In finance, RAG has been explored to improve factual correctness and reduce hallucination:

\begin{itemize}
    \item \citet{shah2024multi} introduce KG\_RAG and RAG\_SEM, two architectures for multi-document financial QA over earnings reports and filings. KG\_RAG incorporates retrieved triples into LLM input, but it assumes a document-centric setup and lacks a mechanism for behavioral personalization or temporal integrity.

    \item FinSRAG~\cite{xiao2025retrieval} adapts RAG to time series forecasting by retrieving historical price patterns similar to current ones. These are then fed to a generation model (StockLLM) for commentary. While it shows the potential of RAG for quantitative forecasting, it does not use KGs or optimize for user preferences.

    \item SURGE~\cite{kang2023knowledge}, though not finance-specific, illustrates how KGs can support dialogue generation by grounding responses in structured knowledge. However, it does not perform multi-step or time-aware subgraph construction.
\end{itemize}

Unlike these approaches, our work integrates KG structure into RAG retrieval, using SPARQL-based multi-step filtering over both personal transaction and market KGs. We enforce temporal cutoffs to ensure that no information leakage occurs, and we serialize the retrieved subgraphs in JSON-LD to preserve structure while fitting within LLM context limits. Most importantly, we align retrieval with user behavior (via PKG) and market signals (via MKG) to produce recommendations that are simultaneously personalized, profitable, and temporally grounded.

\section{Methodology}

% The Multi-Step RAG Pipeline, as outlined in \Cref{fig:pipe}, begins with a user request. Using said request as inspiration, a retrieval LLM is used to select a subset of relevant entities from the user's transaction history KG. The retrieval LLM does not need to be the same as the final inference FLARKO LLM, but we keep them the same in our experiments. This is done by providing the LLM with the request and the set of transaction entities from the user's PKG. The output from the LLM is parsed and used to create a SPARQL query that retrieves all the triples from the KG that involve the selected entities. These triples are combined into a subgraph and included in the request as a prepended system prompt that includes the subgraph in a serialized JSON-LD format.

% This updated request is then used for the market data retrieval step. This step is largely the same as the transaction data retrieval step, except that the Market KG is used instead of the PKG, and there is more context for the LLM to base its entity selection on since it has the PKG subgraph to refer to as well. The important entities in the MKG are the ten-week price summaries, so those are provided to the LLM for selection. The generated subgraph is once again prepended to the request via a system prompt.

% Then, the final prompt is provided to the FLARKO LLM for asset recommendation generation. 

Our framework extends the financial asset recommendation LLM framework, FLARKO~\cite{rag-flarko}, by employing a multi-step RAG pipeline that dynamically constructs a relevant knowledge base for submission to the LLM. This process ensures that the LLM's recommendations are grounded in both personalized user behavior and timely market data, while remaining within the context limit of the LLM by prioritizing the most relevant information.

\subsection{Knowledge Graph Design}

While LLMs excel at processing and generating natural language, they require explicit contextual grounding for structured decision-making tasks to ensure interpretability, consistency, and robustness. To address this, our framework encodes financial context into two distinct KGs: a user's Personal transaction KG (PKG) and a broader Market KG (MKG). These KGs serve as symbolic inputs that anchor the LLM's reasoning to factual, structured data, mitigating common pitfalls like hallucination. By representing user behavior and market signals as interconnected triples serialized into JSON-LD, we provide the LLM with a transparent and controllable context~\cite{spadea2025bursting}, enabling it to reason over complex financial relationships to generate accurate and relevant recommendations.

\begin{itemize}
    \item \textbf{Personal transaction KG (PKG):} This graph encodes an investor's historical transaction behavior, serving as a proxy for their preferences, risk tolerance, and investment patterns. To construct a user's transaction KG, we extract key features from their transaction logs, including the asset's International Securities Identification Number (ISIN), the transaction type (buy/sell), its value, and its timestamp.

    \item \textbf{Market KG (MKG):} This graph captures broader market-level signals and asset characteristics. To manage the volume of raw price data and fit within the LLM's context window, the historical price series are aggregated into \texttt{TenWeekPriceSummary} entities. Each summary encapsulates an asset's performance over a ten-week interval, detailing the period's high, low, average, and end prices. The MKG also includes descriptive asset metadata, such as its category, sector, and industry.
\end{itemize}

\subsection{Multi-Step RAG Pipeline}

The core of our methodology is a sequential, two-stage retrieval pipeline (as illustrated in \Cref{fig:pipe}) that builds compact, relevant subgraphs from the comprehensive KGs to serve as context for the final recommendation task.

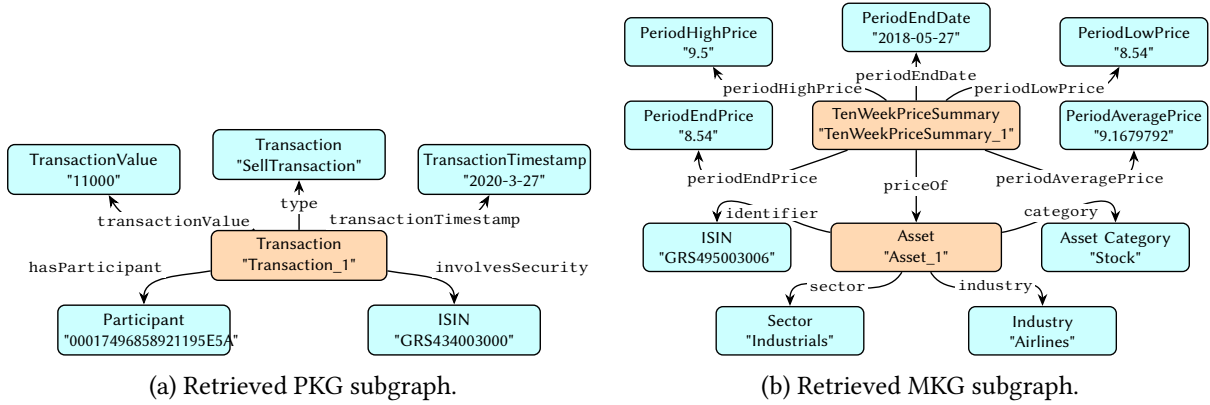
\begin{figure}[h]
    \centering
    \begin{subfigure}{0.49\textwidth}
        \centering
        \resizebox{\linewidth}{!}{%
            % % \documentclass[border=10pt]{standalone}
% % \usepackage{tikz}
% % \usetikzlibrary{arrows.meta, positioning, shapes.geometric}

% \begin{document}

\begin{tikzpicture}[
    node distance=0.7cm and 0.6cm,
    entity/.style={
        rectangle, rounded corners, draw, thick, fill=orange!30,
        minimum width=3.5cm, minimum height=1cm, align=center, font=\sffamily
    },
    attribute/.style={
        rectangle, rounded corners, draw, thick, fill=cyan!20,
        text width=3.2cm, minimum height=1cm, align=center, font=\sffamily
    },
    edge_label/.style={
        fill=white, inner sep=1.5pt, font=\small\ttfamily
    },
    arrow/.style={
        -{Stealth[length=2.5mm, width=2mm]}, thick
    }
]

% === NODES: A CLEAN VERTICAL LAYOUT ===
% Top entity
\node[entity] (Transaction) {Transaction\\"Transaction\_1"};

% Attributes for 'summary', arranged in a clean grid
\node[attribute, left=of Transaction, xshift=1cm, yshift=-1.5cm] (hasParticipant) {Participant\\"00017496858921195E5A"};
\node[attribute, right=of Transaction, xshift=-1cm, yshift=-1.5cm] (involvesSecurity) {ISIN\\"GRS434003000"};
\node[attribute, above left=of Transaction] (TransactionValue) {TransactionValue\\"11000"};
\node[attribute, above right=of Transaction] (TransactionTimestamp) {TransactionTimestamp\\"2020-3-27"};
\node[attribute, above=1cm of Transaction] (type) 
{Transaction \\"SellTransaction"};
% {"BuyTransaction" or\\"SellTransaction"};

% === EDGES: CLEAN AND NON-OVERLAPPING ===
\path[arrow] (Transaction) edge[out=190, in=90] node[edge_label, auto, swap] {hasParticipant} (hasParticipant);
\path[arrow] (Transaction) edge[out=-10, in=90] node[edge_label, auto] {involvesSecurity} (involvesSecurity);
\path[arrow] (Transaction) edge[out=150, in=-60] node[edge_label, pos=0.5] {transactionValue} (TransactionValue);
\path[arrow] (Transaction) edge[out=30, in=-120] node[edge_label, pos=0.5] {transactionTimestamp} (TransactionTimestamp);
\path[arrow] (Transaction) edge node[edge_label] {type} (type);

\end{tikzpicture}

% \end{document}
        }
        \caption{Retrieved PKG subgraph.}
        \label{fig:pkg_example}
    \end{subfigure}
    \hfill % This creates space between the two subfigures
    \begin{subfigure}{0.49\textwidth}
        \centering
        \resizebox{\linewidth}{!}{%
            % % \documentclass[border=10pt]{standalone}
% % \usepackage{tikz}
% % \usetikzlibrary{arrows.meta, positioning, shapes.geometric}

% \begin{document}

% example price summary from dataset: {\"@id\": \"http://securityTrade.com/summary/GRS495003006_2018-05-27\",\"@type\": \"TenWeekPriceSummary\",\"PeriodAveragePrice\": {\"@type\": \"xsd:decimal\",\"@value\": \"9.1679792\"},\"PeriodEndDate\": {\"@type\": \"xsd:date\",\"@value\": \"2018-05-27\"},\"PeriodEndPrice\": {\"@type\": \"xsd:decimal\",\"@value\": \"8.54\"},\"PeriodHighPrice\": {\"@type\": \"xsd:decimal\",\"@value\": \"9.5\"},\"PeriodLowPrice\": {\"@type\": \"xsd:decimal\",\"@value\": \"8.54\"},\"priceOf\": \"urn:isin:GRS495003006\"}

\begin{tikzpicture}[
    node distance=0.7cm and 0.8cm,
    entity/.style={
        rectangle, rounded corners, draw, thick, fill=orange!30,
        minimum width=3.5cm, minimum height=1cm, align=center, font=\sffamily
    },
    attribute/.style={
        rectangle, rounded corners, draw, thick, fill=cyan!20,
        text width=2.8cm, minimum height=1cm, align=center, font=\sffamily
    },
    edge_label/.style={
        fill=white, inner sep=1.5pt, font=\small\ttfamily
    },
    arrow/.style={
        -{Stealth[length=2.5mm, width=2mm]}, thick
    }
]

% === NODES: A CLEAN VERTICAL LAYOUT ===
% Top entity
% \node[entity] (summary) {Ten Week Price Summary};
% OS: I made this into a generic price summary because the 10 week interval could be taken as arbitrary
\node[entity] (summary) {TenWeekPriceSummary\\"TenWeekPriceSummary\_1"};

% Attributes for 'summary', arranged in a clean grid
\node[attribute, left=of summary] (PeriodEndPrice) {PeriodEndPrice\\"8.54"};
\node[attribute, right=of summary] (PeriodAvgPrice) {PeriodAveragePrice\\"9.1679792"};
\node[attribute, above=of PeriodEndPrice] (PeriodHighPrice) {PeriodHighPrice\\"9.5"};
\node[attribute, above=of PeriodAvgPrice] (PeriodLowPrice) {PeriodLowPrice\\"8.54"};
\node[attribute, above=1cm of summary] (PeriodEndDate) {PeriodEndDate\\"2018-05-27"};

% Bottom entity
\node[entity, below=1.5cm of summary] (asset) {Asset\\"Asset\_1"};

% Attributes for 'asset', arranged in a clean grid
\node[attribute, left=of asset] (identifier) {ISIN\\"GRS495003006"};
\node[attribute, right=of asset] (category) {Asset Category\\"Stock"};
% --- ADJUSTED LINES BELOW ---
\node[attribute, below=of identifier, xshift=1.5cm] (sector) {Sector\\"Industrials"};
\node[attribute, below=of category, xshift=-1.5cm] (industry) {Industry\\"Airlines"};

% === EDGES: CLEAN AND NON-OVERLAPPING ===
% Central connection
\path[arrow] (summary) edge node[edge_label, swap] {priceOf} (asset);

% Edges for 'Summary'
\path[arrow] (summary) edge[out=-160, in=-90] node[edge_label] {periodEndPrice} (PeriodEndPrice);
\path[arrow] (summary) edge[out=-20, in=-90] node[edge_label, swap] {periodAveragePrice} (PeriodAvgPrice);
\path[arrow] (summary) edge[out=140, in=-60] node[edge_label, pos=0.5] {periodHighPrice} (PeriodHighPrice);
\path[arrow] (summary) edge[out=40, in=-120] node[edge_label, swap, pos=0.5] {periodLowPrice} (PeriodLowPrice);
\path[arrow] (summary) edge node[edge_label] {periodEndDate} (PeriodEndDate);

% Edges for 'Asset'
\path[arrow] (asset) edge[out=170, in=90] node[edge_label, pos=0.4] {identifier} (identifier);
\path[arrow] (asset) edge[out=10, in=90] node[edge_label, pos=0.4] {category} (category);
\path[arrow] (asset) edge[out=-120, in=90] node[edge_label] {sector} (sector);
\path[arrow] (asset) edge[out=-60, in=90] node[edge_label, swap] {industry} (industry);

\end{tikzpicture}

% \end{document}
        }
        \caption{Retrieved MKG subgraph.}
        \label{fig:mkg_example}
    \end{subfigure}
    \caption{Examples of retrieved subgraphs from the (a) Personal transaction KG and (b) Market KG, each centered on a selected entity.}
    \label{fig:retrieved_examples}
\end{figure}

\begin{enumerate}
    \item \textbf{Personal Transaction Retrieval (PTR):} The process begins with an initial user request. A retrieval-focused LLM is provided with this request and the set of all transaction entities from the user's PKG. This LLM selects a subset of entities deemed most behaviorally relevant, i.e., aligned with the user's past transaction history. From this selection, a SPARQL \texttt{CONSTRUCT} query is programmatically generated using the SPARQL query template given below, where \texttt{NODE\_LIST} is a space-delimited list of the selected entities.
    % in \Cref{lst:sparql_query}. 
    This query retrieves a subgraph containing all triples where the selected entities appear as either the subject or the object, as shown in \Cref{fig:pkg_example}. The resulting subgraph, representing the most salient aspects of the user's transaction history, is serialized and prepended to the original request via a system prompt.

    \item \textbf{Market Retrieval (MR):} The updated request, now containing the retrieved PKG subgraph, proceeds to the market retrieval stage. A retrieval LLM is presented with this enriched context and the set of \texttt{TenWeekPriceSummary} entities from the MKG. The LLM selects relevant market entities, and the same SPARQL query template shown below 
    % (\Cref{lst:sparql_query}) 
    is used to retrieve the corresponding market subgraph, as shown in \Cref{fig:mkg_example}. This market-focused subgraph is also serialized and prepended to the request via another system prompt.

    \item \textbf{Recommendation Generation:} The final fully contextualized prompt, containing both the retrieved PKG and Market KG subgraphs, is passed to the FLARKO LLM. Grounded in this tailored, dual-faceted context, the model generates the final list of asset recommendations.
\end{enumerate}

\begin{tcolorbox}[title=SPARQL Query Template, coltitle=white, fonttitle=\bfseries, sharp corners=southwest, enhanced, label={lst:sparql_query}]
\ttfamily
% OS: where is this namespace being used? I suppose it is used for the node_list.
% PREFIX ns: <http://securityTrade.com/ns\#>\\
CONSTRUCT \{ ?s ?p ?o \}\\
WHERE \{\\
\verb|    | VALUES ?node \{ [NODE\_LIST] \}\\
\verb|    | \{ ?node ?p ?o . BIND(?node as ?s) \}\\
\verb|    | UNION\\
\verb|    | \{ ?s ?p ?node . BIND(?node as ?o) \}\\
\}
\end{tcolorbox}

% \begin{figure}[h]
%     \centering
%     \resizebox{\columnwidth}{!}{%
%         \input{figures/pkg-schema}
%     }
%     \caption{Example of a retrieved subgraph from the Personal transaction KG (PKG), centered on a selected transaction entity.
%     }
%     \label{fig:pkg_example}
% \end{figure}

% \begin{figure}[h]
%     \centering
%     \resizebox{\columnwidth}{!}{%
%         \input{figures/mkg-schema}
%     }
%     \caption{Example of a retrieved subgraph from the Market KG (MKG), centered on a selected \texttt{TenWeekPriceSummary} entity.}
%     \label{fig:mkg_example}
% \end{figure}

\subsection{Dataset}
We evaluate our framework using the \textbf{FAR-Trans dataset}~\cite{sanz2024far}, a real-world collection of anonymized financial data containing customer transaction histories, asset price data, and investor profile information. Our testing period spans from December 1, 2021, to November 29, 2022. Within this period, test instances are generated every two weeks, with each instance using the corresponding date as its \texttt{RECOMMENDATION\_DATE}. This setup allows us to simulate a realistic scenario where recommendations are made periodically based on evolving historical data. A critical aspect of our approach is the enforcement of temporal consistency. For each recommendation request, the KGs are built using only information available before \texttt{RECOMMENDATION\_DATE}. This strict cutoff prevents data leakage and ensures that the model's recommendations are based on a historically accurate snapshot of information.

A concern with using pre-trained models like Qwen3 is the potential for data leakage, where the model may have been trained on the evaluation dataset. However, the pre-training process for Qwen3 involved sourcing data primarily from general web crawls, PDF-like documents, and synthetic data generation focused on domains such as mathematics and coding~\cite{Team_2025}. 
% OS: TODO add a ref here
% FS: Done
This methodology suggests it is unlikely that a specialized dataset like FAR-Trans was explicitly included in the training corpus. Furthermore, even in the event of some data overlap, the nature of our task, which grounds recommendations in the specific user and market KGs provided in the prompt, significantly mitigates the impact of any potential data leakage.

\section{Evaluation}

\subsection{Evaluation Metrics}
To measure the quality of the recommendations, we use three variants of the \textbf{Hits@3} metric, which assesses the top three assets recommended by the model. Each metric evaluates the recommendations against outcomes in the 180-day period following the \texttt{RECOMMENDATION\_DATE}.

\begin{itemize}
    \item \textbf{Pref@3 (Preference Alignment):} This measures the hit rate of the recommendations against the set of assets that the user \textit{actually purchased} during the subsequent 180-day window. It quantifies how well the model aligns with the user's revealed preferences.

    \item \textbf{Prof@3 (Profitability):} This measures the hit rate against assets that yielded a \textit{positive financial return} over the same 180-day period. It assesses the financial soundness of the recommendations.

    \item \textbf{Comb@3 (Combined Score):} This is our primary metric for success, as it measures the hit rate against the intersection of the two sets above, assets that were \textit{both purchased by the user and were profitable}. A high \texttt{Comb@3} score indicates that the model is generating actionable, high-quality advice that is both behaviorally aligned and financially beneficial.
\end{itemize}

\subsection{Baselines and Model Variants}

We compare the performance of RAG-FLARKO against:

\begin{enumerate}
    \item \textbf{FLARKO Baseline:} The original FLARKO framework, which injects both the PKG and MKG directly into the LLM context without any intermediate retrieval steps. This baseline highlights the limitations of full KG injection in terms of context efficiency and scalability.
    
    \item \textbf{Parallel RAG-FLARKO:} A baseline retrieval-augmented variant that applies the same two-stage pipeline but disables inter-stage context propagation. Specifically, the MR step operates without access to the context retrieved during the PTR stage. This isolates the impact of context chaining across retrieval stages.
    
    \item \textbf{Multi-stage RAG-FLARKO:} Our full model, which performs sequential retrieval with context propagation from PTR to MR. This version enables the MR stage to incorporate behaviorally relevant information extracted from the PTR step, resulting in more targeted market signal retrieval and improved final recommendations.

\end{enumerate}
 
% OS: attempting to rewrite the following with a slightly better organization

% \subsection{Evaluated Models}
We test all variants using the Qwen3-0.6B and Qwen3-1.7B LLMs~\cite{huggingface2024qwen3} to highlight the benefits of context-efficient retrieval under limited model capacity.
% OS: I suppose we let readers look at the chart and decide for themselves.
% , which previously underperformed in the original FLARKO framework due to their limited context capacities. 
% OS: rephrased as below
% The optimizations of our Multi-step RAG framework are ideal for improving the results of these two small models, as it optimization context length usage.
% OS: this is redundant now. I am moving this to the conclusion.
% Our multi-step RAG approach is specifically designed to optimize context usage by retrieving and injecting only the most relevant knowledge subgraphs, making it particularly well-suited to enhancing the performance of these compact models.
% We include comparisons against the original FLARKO baseline to evaluate the overall improvement from introducing multi-step retrieval.

% OS: incorporated this into the list above
% \subsection{Ablation Study}
% % In addition to comparing our RAG-FLARKO results against the base FLARKO results, 
% To isolate the contribution of cross-stage contextual grounding, we conduct an ablation study to compare our multi-step RAG-FLARKO framework against a parallel framework where the MR step would not have access to the retrieved context from the PTR step to demonstrate the benefits of this extra context.
% By removing this behavioral context, we assess the impact of user-specific signal propagation on the quality of retrieved market information and downstream recommendations. 

\section{Results}

\begin{figure}[t]
    \centering
    \includegraphics[width=\linewidth]{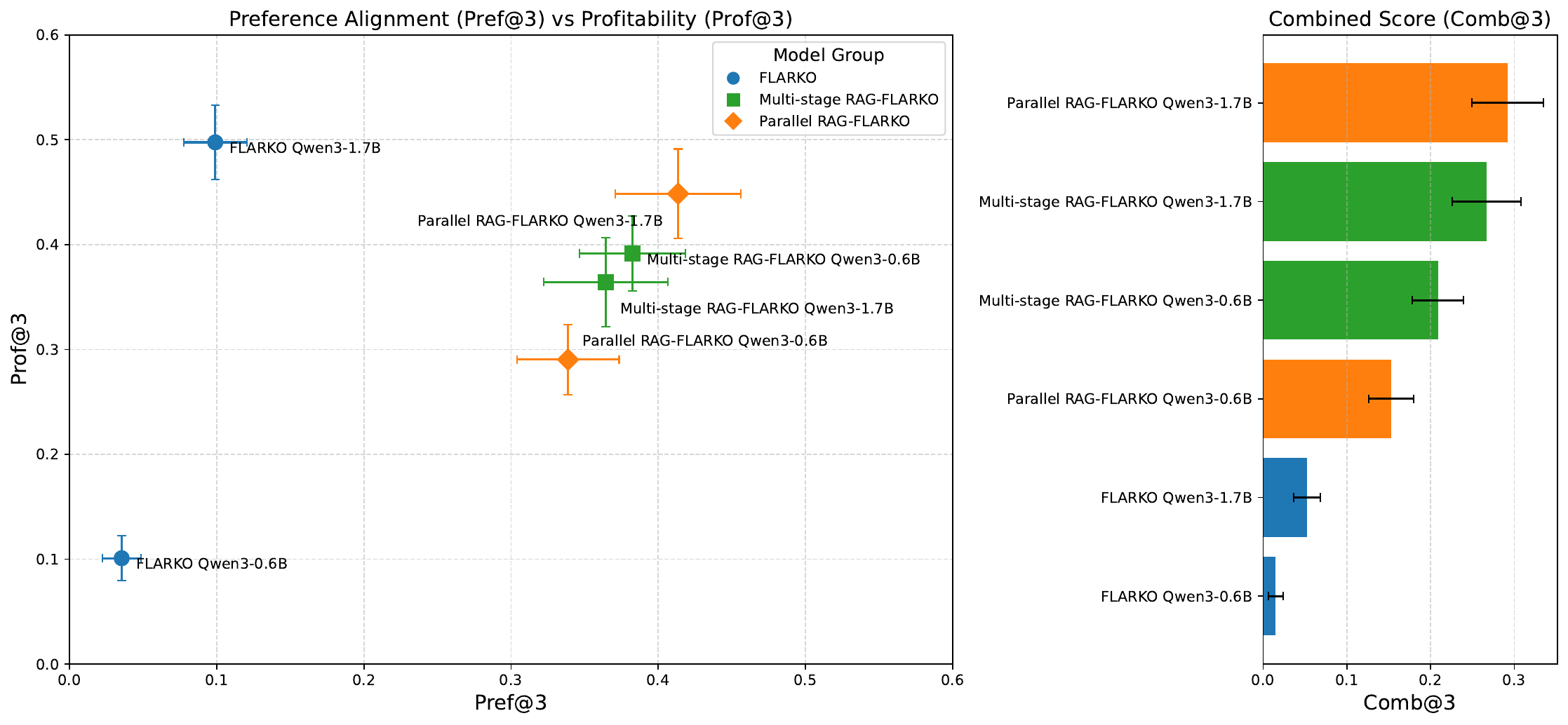}
    % \captionsetup{justification=centering} 
    \caption{Performance Comparison of FLARKO, RAG-FLARKO, and Parallel RAG-FLARKO \\
\normalfont\small 
The left panel plots preference alignment (Pref@3) against profitability (Prof@3) for all models. The right panel reports Comb@3, which quantifies how often the model recommends assets that are both profitable and behaviorally aligned. The Standard Error of Proportions are plotted in the graphs as  well.}
    \label{fig:chart}
\end{figure}

% OS: let's tone it down a bit ;)
% \paragraph{Superiority of RAG-FLARKO over Baseline}
\paragraph{Overall Performance Gains:}
The empirical results, presented in \Cref{fig:chart}, demonstrate the significant advantages of the RAG-FLARKO framework over the baseline FLARKO models. For both the Qwen3-0.6B and Qwen3-1.7B models, the RAG-FLARKO (both parallel and multi-stage) implementations yield substantial improvements across almost all key metrics. Most notably, the right panel shows a dramatic increase in the \textbf{Comb@3} score, our primary metric for evaluating recommendations that are simultaneously profitable and aligned with user preferences.

This performance gain underscores the efficacy of our retrieval-augmented approach. The baseline FLARKO models, which fill the entire context with the KGs, struggled due to the limited context capacities of the smaller Qwen3 models. 
In contrast, RAG-FLARKO's retrieval pipeline constructs compact and highly relevant subgraphs, enabling more efficient use of the available context window. This targeted approach allows the LLM to ground its reasoning in the most salient user and market data, leading to higher-quality recommendations.

\paragraph{Impact of Inter-Stage Context:}
% \paragraph{Benefits of Sequential, Multi-Step Retrieval}
Our comparison between the multi-stage RAG-FLARKO pipeline and a parallel variant, confirms the value of contextual chaining between retrieval stages in the smaller Qwen3\_0.6B. As shown in \Cref{fig:chart}, the multi-stage Qwen3\_0.6B model consistently outperform its parallel counterpart. 

The parallel RAG framework, which retrieves personal and market information independently, lacks the user-specific context during market signal selection.
The multi-stage pipeline, however, uses the output of the PTR stage to inform the MR stage, which ensures that the retrieved market data from the MKG is conditioned on the user's unique behavioral patterns in the PKG. The resulting improvement in all metrics highlights that this inter-stage context is crucial for identifying financial signals that are not just broadly relevant, but specifically tailored to the individual investor, thereby enhancing both profitability and preference alignment.

\paragraph{Impact of Model Size:}

Interestingly, the benefits of the inter-stage context do not apply to the larger Qwen3\_1.7B model, which does not improve with the multi-stage pipeline. This suggests that larger models may possess sufficient reasoning capacity to handle entity selection without this additional context, whereas smaller models require that explicit guidance to perform optimally.

The observation that the multi-stage RAG pipeline disproportionately benefits the smaller Qwen3-0.6B model is a critical finding, supporting the feasibility of deploying RAG-FLARKO in real-world, resource-constrained environments. While the addition of inter-stage context did not significantly help the larger Qwen3-1.7B model, it transformed the Qwen3-0.6B model from the poorest performer into a highly competitive one. This result is particularly relevant to the deployment considerations discussed in \Cref{sec:discussion}, where smaller, quantized models are ideal for execution on edge servers or client devices to preserve privacy and reduce latency. The success of the enhanced smaller model demonstrates a practical pathway for creating effective financial recommendation systems that do not rely on large, resource-intensive models, thereby making advanced, personalized AI tools more scalable and accessible. This makes the multi-stage pipeline a valuable approach in federated learning environments, where smaller, more efficient models are often necessary due to resource constraints on client devices.

\section{Discussion}
\label{sec:discussion}

\paragraph{Rationale for Multi-Step Retrieval:} While our multi-step retrieval process incurs an additional LLM call over just having a single-step process (in which PKG and MKG are merged and queried jointly), this design yields important benefits. 
% However, there are several significant drawbacks to the single-step process over the multi-step process. 
The single-step process introduces several limitations that negatively impact both retrieval quality and model interpretability.
Most critically, it reduces the contextual information for the LLM to select its list of relevant entities. In contrast, the multi-step process enables the MR stage to condition its entity selection on behavioral context retrieved during the PTR stage. 
% OS: rephrased the following sentence as above as it's better to be explict about what the first and retrieval stages are
% is that the second retrieval from the market KG can use context from the first retrieval to make a better selection of entities. 
The context retrieved in the PTR stage includes a lot of important information from the user's PKG that informs the LLM of the user's specific preferences, so it can focus on the relevant entities. On the other hand, a single step would have the LLM select relevant entities for both the PKG and MKG without the previously retrieved context from within the PKG, since it would be working off only the list of entities. 
Additionally, it would also make the task much harder as the LLM would have to reason over the combined list of entities all at once rather than separately, overwhelming the LLM due to the PKG and MKG's entities representing different types of information.

% OS: add something about the ten week summary of the MKG, and what might happen if we don't have that summarization
% \paragraph{Choice of Ten-Week Summaries:} 
\paragraph{MKG Summaries:} 
The market graph is massive, and fitting it in an LLM's context window is infeasible.
While our RAG framework alleviates this issue through selective retrieval, the use of \texttt{TenWeekPriceSummary} entities in the MKG further improves efficiency and interpretability.
% OS: rephrased as above
% Of course, our RAG framework is designed to combat that problem, but the summaries still help regardless. 
Individual daily data points, though granular, take up almost as much space as a summary in the model's context window. With summaries, we can provide much more information, over a much longer time span, to the LLM than we can with daily data points. Additionally, it is much more effective to generalize patterns with macro data than with micro data.

% OS: @FS, please check if you agree with this modularity argument
\paragraph{Deployment Considerations:}
The modular architecture of the RAG-FLARKO pipeline lends itself well to deployment in resource-constrained or privacy-sensitive environments. The two retrieval stages (PTR and MR) operate over structured KGs and can be executed locally on client devices or edge servers. This design allows users to retain control over sensitive transaction data and minimizes the need to transmit personal information to centralized infrastructure.

While our current implementation uses the same LLM for all three stages, a natural future optimization would be to substitute a lightweight model for the PTR component. Because PTR focuses solely on selecting behaviorally relevant entities from the user’s PKG, it could be effectively handled by a smaller LLM that has been quantized for edge deployment. This would reduce memory and compute requirements, enabling real-time inference on mobile or embedded devices.

Only the final recommendation generation step requires access to a full-scale LLM, which can be hosted in a secure cloud environment or exposed via a privacy-preserving API. This separation of concerns enables hybrid deployment strategies that reduce latency, improve scalability, and uphold data sovereignty.

Emerging techniques such as model quantization and distillation can further enhance RAG-FLARKO's deployability. Quantized models can dramatically reduce memory footprint and inference latency with minimal performance loss, making them ideal for on-device execution. Cross-device inference approaches could also be explored, where retrieval runs on the user’s device while generation is offloaded to a remote server, maintaining a balance between responsiveness and resource efficiency. For example, in mobile-based financial advisory systems, users could locally construct personalized subgraphs and perform retrieval in real time, sending only distilled, behavior- and market-relevant context to a centralized LLM.

\section{Conclusion}

RAG has emerged as a promising approach to tackle the knowledge-intensiveness of financial AI tasks.
This is particularly important in finance, where pretrained models may lack access to up-to-date information or personalized behavioral insights needed for trustworthy recommendations.
% and it is crucial in finance for overcoming the limitations of static trained models (which might be outdated or unaware of niche personalized details for effective recommendations).

Our RAG-FLARKO framework introduces a multi-step retrieval pipeline designed to efficiently navigate and inject only the most relevant knowledge subgraphs into the LLM’s context. This architecture is optimized for compact models, addressing context limitations while maintaining personalization and factual grounding.
% OS: moved from the Evaluation section
% OS: rephrased above
% Our multi-step RAG approach is specifically designed to optimize context usage by retrieving and injecting only the most relevant knowledge subgraphs, making it particularly well-suited to enhancing the performance of these compact models.
We include empirical comparisons against the original FLARKO baseline to evaluate the overall improvement from introducing multi-step and parallel retrieval.

% OS: TODO add a summary of our results and sing some praise about our methods
% OS: I don't like to over use "superior"
Our empirical evaluation validates this promise, demonstrating that the RAG-FLARKO framework delivers substantial gains in recommendation quality. The results show that our method successfully generates recommendations that are both profitable and aligned with user behavior, as evidenced by consistent improvement in Comb@3 scores. Crucially, our ablation study revealed that the benefits of the sequential, multi-stage retrieval process are most pronounced for smaller, more resource-constrained models. This confirms that a structured, context-aware retrieval pipeline can effectively overcome the inherent limitations of smaller LLMs, elevating their performance to be competitive with much larger models. This finding provides a clear pathway for developing financially-grounded, personalized AI systems that are not only effective but also efficient and practical for real-world deployment.

% OS: future directions -- potentially useful for the CRAFT proposal, and more neurosymbolic integration. @FS, please check
Future work could explore integrating symbolic reasoning modules to assist the reasoning tasks currently handled by the LLM. For example, symbolic reasoning engines could enforce domain-specific constraints (e.g., regulatory rules, portfolio diversification requirements) during subgraph construction, ensuring that generated recommendations are not only grounded but are also compliant. 
Additionally, we could leverage inferred or declared user characteristics, such as customer type (mass, premium, legal entity, professional), the investors' risk level (conservative, moderate, aggressive), and their investment capacity (large-scale, medium-scale, small-scale) to to further personalize both the retrieval and generation stages of the pipeline. For instance, a risk-averse investor might benefit from a PTR stage that prioritizes historically stable assets and an MR stage that emphasizes downside protection signals (e.g., volatility indexes, bond yields). Conversely, an aggressive investor's prompts could steer the pipeline toward high-growth sectors, momentum signals, or emerging asset classes. This adaptive prompting would allow the LLM to tailor both its interpretation of user behavior and its generation of recommendations, effectively embedding a user’s financial persona into the decision-making process.

%% The declaration on generative AI comes in effect
%% in Janary 2025. See also
%% https://ceur-ws.org/GenAI/Policy.html
\section*{Declaration on Generative AI}
 During the preparation of this work, the authors used ChatGPT, Gemini and Grammarly in order to rephrase some of the sentences and also to fix grammar and spelling issues. After using these tools and services, the authors reviewed and edited the content as needed and take full responsibility for the publication’s content. 

\section{Resource Contributions}

All research artifacts, including code and documentation, are released under the MIT license. To support transparency and reproducibility, we maintain an open-source GitHub repository~\cite{rag-flarko} containing all software artifacts. The specific version of the code used in this paper is available under the tag \texttt{RAGE-KG\_2025} at \url{https://github.com/brains-group/FLARKO/releases/tag/RAGE-KG_2025}.

%%
%% Define the bibliography file to be used
\bibliography{references}

\begin{thebibliography}{18}
\expandafter\ifx\csname natexlab\endcsname\relax\def\natexlab#1{#1}\fi
\providecommand{\url}[1]{\texttt{#1}}
\providecommand{\href}[2]{#2}
\providecommand{\path}[1]{#1}
\providecommand{\DOIprefix}{doi:}
\providecommand{\ArXivprefix}{arXiv:}
\providecommand{\URLprefix}{URL: }
\providecommand{\Pubmedprefix}{pmid:}
\providecommand{\doi}[1]{\href{http://dx.doi.org/#1}{\path{#1}}}
\providecommand{\Pubmed}[1]{\href{pmid:#1}{\path{#1}}}
\providecommand{\bibinfo}[2]{#2}
\ifx\xfnm\relax \def\xfnm[#1]{\unskip,\space#1}\fi
%Type = Misc
\bibitem[{Spadea and Seneviratne(2025)}]{rag-flarko}
\bibinfo{author}{F.~Spadea}, \bibinfo{author}{O.~Seneviratne}, \bibinfo{title}{{FLARKO: Financial Language-model for Asset Recommendation with Knowledge-graph Optimization}}, \bibinfo{howpublished}{\url{https://github.com/brains-group/FLARKO/tree/RAG}}, \bibinfo{year}{2025}.
%Type = Misc
\bibitem[{Ethayarajh et~al.(2024)Ethayarajh, Xu, Muennighoff, Jurafsky, and Kiela}]{ethayarajh2024kto}
\bibinfo{author}{K.~Ethayarajh}, \bibinfo{author}{W.~Xu}, \bibinfo{author}{N.~Muennighoff}, \bibinfo{author}{D.~Jurafsky}, \bibinfo{author}{D.~Kiela}, \bibinfo{title}{Kto: Model alignment as prospect theoretic optimization}, \bibinfo{year}{2024}.
%Type = Inproceedings
\bibitem[{Spadea and Seneviratne(2025)}]{spadea2025federated}
\bibinfo{author}{F.~Spadea}, \bibinfo{author}{O.~Seneviratne},
\newblock \bibinfo{title}{Federated fine-tuning of large language models: Kahneman-tversky vs. direct preference optimization},
\newblock in: \bibinfo{booktitle}{Companion Proceedings of the ACM on Web Conference 2025}, \bibinfo{year}{2025}, pp. \bibinfo{pages}{1757--1760}.
%Type = Article
\bibitem[{Sanz-Cruzado et~al.(2024)Sanz-Cruzado, Droukas, and McCreadie}]{sanz2024far}
\bibinfo{author}{J.~Sanz-Cruzado}, \bibinfo{author}{N.~Droukas}, \bibinfo{author}{R.~McCreadie},
\newblock \bibinfo{title}{Far-trans: An investment dataset for financial asset recommendation},
\newblock \bibinfo{journal}{arXiv preprint arXiv:2407.08692}  (\bibinfo{year}{2024}).
%Type = Article
\bibitem[{Guo et~al.(2020)Guo, Zhuang, Qin, Zhu, Xie, Xiong, and He}]{guo2020survey}
\bibinfo{author}{Q.~Guo}, \bibinfo{author}{F.~Zhuang}, \bibinfo{author}{C.~Qin}, \bibinfo{author}{H.~Zhu}, \bibinfo{author}{X.~Xie}, \bibinfo{author}{H.~Xiong}, \bibinfo{author}{Q.~He},
\newblock \bibinfo{title}{A survey on knowledge graph-based recommender systems},
\newblock \bibinfo{journal}{IEEE Transactions on Knowledge and Data Engineering} \bibinfo{volume}{34} (\bibinfo{year}{2020}) \bibinfo{pages}{3549--3568}.
%Type = Article
\bibitem[{Sun et~al.(2025)Sun, Pan, Qi, and Gao}]{sun2025knowledge}
\bibinfo{author}{S.~Sun}, \bibinfo{author}{X.~Pan}, \bibinfo{author}{S.~Qi}, \bibinfo{author}{J.~Gao},
\newblock \bibinfo{title}{Knowledge enhanced prompt learning framework for financial news recommendation},
\newblock \bibinfo{journal}{Pattern Recognition} \bibinfo{volume}{163} (\bibinfo{year}{2025}) \bibinfo{pages}{111461}.
%Type = Article
\bibitem[{Shen et~al.(2025)Shen, Wang, Chew, Hu, and Wang}]{shen2025artificial}
\bibinfo{author}{Z.~Shen}, \bibinfo{author}{Z.~Wang}, \bibinfo{author}{J.~Chew}, \bibinfo{author}{K.~Hu}, \bibinfo{author}{Y.~Wang},
\newblock \bibinfo{title}{Artificial intelligence empowering robo-advisors: A data-driven wealth management model analysis},
\newblock \bibinfo{journal}{International Journal of Management Science Research} \bibinfo{volume}{8} (\bibinfo{year}{2025}) \bibinfo{pages}{1--12}.
%Type = Article
\bibitem[{Ren et~al.(2019)Ren, Long, and Xu}]{ren2019financial}
\bibinfo{author}{J.~Ren}, \bibinfo{author}{J.~Long}, \bibinfo{author}{Z.~Xu},
\newblock \bibinfo{title}{Financial news recommendation based on graph embeddings},
\newblock \bibinfo{journal}{Decision Support Systems} \bibinfo{volume}{125} (\bibinfo{year}{2019}) \bibinfo{pages}{113115}.
%Type = Inproceedings
\bibitem[{Li and Sanna~Passino(2024)}]{li2024findkg}
\bibinfo{author}{X.~V. Li}, \bibinfo{author}{F.~Sanna~Passino},
\newblock \bibinfo{title}{Findkg: Dynamic knowledge graphs with large language models for detecting global trends in financial markets},
\newblock in: \bibinfo{booktitle}{Proceedings of the 5th ACM international conference on AI in finance}, \bibinfo{year}{2024}, pp. \bibinfo{pages}{573--581}.
%Type = Inproceedings
\bibitem[{Tang et~al.(2023)Tang, Zhao, and Yu}]{tang2023intelligent}
\bibinfo{author}{C.~M. Tang}, \bibinfo{author}{Y.~Zhao}, \bibinfo{author}{X.~Yu},
\newblock \bibinfo{title}{Intelligent stock recommendation system based on generalized financial knowledge graph},
\newblock in: \bibinfo{booktitle}{Third International Conference on Intelligent Computing and Human-Computer Interaction (ICHCI 2022)}, volume \bibinfo{volume}{12509}, \bibinfo{organization}{SPIE}, \bibinfo{year}{2023}, pp. \bibinfo{pages}{332--338}.
%Type = Article
\bibitem[{Verma et~al.(2023)Verma, Sengupta, Simanta, Chen, Perge, Pillai, McCrae, and Buitelaar}]{verma2023empowering}
\bibinfo{author}{G.~Verma}, \bibinfo{author}{S.~Sengupta}, \bibinfo{author}{S.~Simanta}, \bibinfo{author}{H.~Chen}, \bibinfo{author}{J.~A. Perge}, \bibinfo{author}{D.~Pillai}, \bibinfo{author}{J.~P. McCrae}, \bibinfo{author}{P.~Buitelaar},
\newblock \bibinfo{title}{Empowering recommender systems using automatically generated knowledge graphs and reinforcement learning},
\newblock \bibinfo{journal}{arXiv preprint arXiv:2307.04996}  (\bibinfo{year}{2023}).
%Type = Article
\bibitem[{Lewis et~al.(2020)Lewis, Perez, Piktus, Petroni, Karpukhin, Goyal, K{\"u}ttler, Lewis, Yih, Rockt{\"a}schel et~al.}]{lewis2020retrieval}
\bibinfo{author}{P.~Lewis}, \bibinfo{author}{E.~Perez}, \bibinfo{author}{A.~Piktus}, \bibinfo{author}{F.~Petroni}, \bibinfo{author}{V.~Karpukhin}, \bibinfo{author}{N.~Goyal}, \bibinfo{author}{H.~K{\"u}ttler}, \bibinfo{author}{M.~Lewis}, \bibinfo{author}{W.-t. Yih}, \bibinfo{author}{T.~Rockt{\"a}schel}, et~al.,
\newblock \bibinfo{title}{Retrieval-augmented generation for knowledge-intensive nlp tasks},
\newblock \bibinfo{journal}{Advances in neural information processing systems} \bibinfo{volume}{33} (\bibinfo{year}{2020}) \bibinfo{pages}{9459--9474}.
%Type = Article
\bibitem[{Shah et~al.(2024)Shah, Ryali, and Venkatesh}]{shah2024multi}
\bibinfo{author}{S.~Shah}, \bibinfo{author}{S.~Ryali}, \bibinfo{author}{R.~Venkatesh},
\newblock \bibinfo{title}{Multi-document financial question answering using llms},
\newblock \bibinfo{journal}{CoRR}  (\bibinfo{year}{2024}).
%Type = Article
\bibitem[{Xiao et~al.(2025)Xiao, Jiang, Qian, Chen, He, Xu, Jiang, Li, Weng, Peng et~al.}]{xiao2025retrieval}
\bibinfo{author}{M.~Xiao}, \bibinfo{author}{Z.~Jiang}, \bibinfo{author}{L.~Qian}, \bibinfo{author}{Z.~Chen}, \bibinfo{author}{Y.~He}, \bibinfo{author}{Y.~Xu}, \bibinfo{author}{Y.~Jiang}, \bibinfo{author}{D.~Li}, \bibinfo{author}{R.-L. Weng}, \bibinfo{author}{M.~Peng}, et~al.,
\newblock \bibinfo{title}{Retrieval-augmented large language models for financial time series forecasting},
\newblock \bibinfo{journal}{arXiv preprint arXiv:2502.05878}  (\bibinfo{year}{2025}).
%Type = Article
\bibitem[{Kang et~al.(2023)Kang, Kwak, Baek, and Hwang}]{kang2023knowledge}
\bibinfo{author}{M.~Kang}, \bibinfo{author}{J.~M. Kwak}, \bibinfo{author}{J.~Baek}, \bibinfo{author}{S.~J. Hwang},
\newblock \bibinfo{title}{Knowledge graph-augmented language models for knowledge-grounded dialogue generation},
\newblock \bibinfo{journal}{arXiv preprint arXiv:2305.18846}  (\bibinfo{year}{2023}).
%Type = Inproceedings
\bibitem[{Spadea and Seneviratne(2025)}]{spadea2025bursting}
\bibinfo{author}{F.~Spadea}, \bibinfo{author}{O.~Seneviratne},
\newblock \bibinfo{title}{Bursting the filter bubble with knowledge graph inversion},
\newblock in: \bibinfo{booktitle}{Companion Publication of the 17th ACM Web Science Conference 2025}, \bibinfo{year}{2025}, pp. \bibinfo{pages}{39--43}.
%Type = Misc
\bibitem[{{Qwen Team}(2025)}]{Team_2025}
\bibinfo{author}{{Qwen Team}}, \bibinfo{title}{{Qwen3: Think deeper, act faster}}, \bibinfo{year}{2025}. \URLprefix \url{https://qwenlm.github.io/blog/qwen3/}.
%Type = Misc
\bibitem[{{Hugging Face}(2024)}]{huggingface2024qwen3}
\bibinfo{author}{{Hugging Face}}, \bibinfo{title}{Qwen3-0.6b}, \bibinfo{howpublished}{\url{https://huggingface.co/Qwen/Qwen3-0.6B}}, \bibinfo{year}{2024}.

\end{thebibliography}

\end{document}